\documentclass{article}
\usepackage{ijcai23}

\usepackage{times}
\usepackage{soul}
\usepackage{url}
\usepackage[hidelinks]{hyperref}
\usepackage[utf8]{inputenc}
\usepackage[small]{caption}
\usepackage{graphicx}
\usepackage{amsmath}
\usepackage{amsthm}
\usepackage{booktabs}
\usepackage{algorithm}
\usepackage{algorithmic}
\usepackage[switch]{lineno}

\usepackage{bm}
\usepackage{amssymb}
\usepackage{mathrsfs}

\title{An Effective and Efficient Time-aware Entity Alignment Framework via Two-aspect Three-view Label Propagation}

\author{
Li Cai$^{1,2}$
\and
Xin Mao$^1$\and
Youshao Xiao$^5$\and
Changxu Wu$^6$\And
Man Lan$^{1,3,4}$\thanks{Corresponding author}
\affiliations
$^1$School of Computer Science and Technology, East China Normal University, Shanghai 200062, China\\
$^2$College of Computer Science and Technology, Guizhou University, Guiyang 5220025, China\\
$^3$Shanghai Institute of AI for Education, East China Normal University\\
$^4$Lingang Laboratory, Shanghai 200031, China\\
$^5$Ant Group, China\\
$^6$Department of Industrial Engineering, Tsinghua University\\
\emails
lcai2020@stu.ecnu.edu.cn,
xmao@stu.ecnu.edu.cn,
youshao.xys@antgroup.com,
wuchangxu@tsinghua.edu.cn,
mlan@cs.ecnu.edu.cn
}
\begin{document}

\maketitle

\begin{abstract}
Entity alignment (EA) aims to find the equivalent entity pairs between different knowledge graphs (KGs), which is crucial to promote knowledge fusion. With the wide use of temporal knowledge graphs (TKGs), time-aware EA (TEA) methods appear to enhance EA. Existing TEA models are based on Graph Neural Networks (GNN) and achieve state-of-the-art (SOTA) performance, but it is difficult to transfer them to large-scale TKGs due to the scalability issue of GNN. In this paper, we propose an effective and efficient non-neural EA framework between TKGs, namely LightTEA, which consists of four essential components: (1) Two-aspect Three-view Label Propagation, (2) Sparse Similarity with Temporal Constraints, (3) Sinkhorn Operator, and (4) Temporal Iterative Learning. All of these modules work together to improve the performance of EA while reducing the time consumption of the model. Extensive experiments on public datasets indicate that our proposed model significantly outperforms the SOTA methods for EA between TKGs, and the time consumed by LightTEA is only dozens of seconds at most, no more than 10$\%$ of the most efficient TEA method.
\end{abstract}

\section{Introduction}
Knowledge graphs (KGs) describe the real world with structured facts. A fact consists of a head entity, a tail entity, and a relation connecting them, which can be formally represented as a triple $(e_h, r, e_t)$, such as (\emph{George Porter, hasWonPrize, Nobel Prize in Chemistry}). KGs have been widely used in information retrieval~\cite{DBLP:conf/sigir/DietzKM18-InformationRetrieval}, question answering~\cite{DBLP:conf/ijcai/LanHJ0ZW21-QuestionAnswering}, and recommedation systems~\cite{DBLP:journals/tkde/GuoZQZXXH22-RecommenderSystem}. 

Existing KGs ignore the temporal information which indicates when a fact occurred. In the real world, some facts only happened at a specific time. Therefore, Wikidata~\cite{DBLP:journals/cacm/VrandecicK14-Wikidata} and YOGO2~\cite{DBLP:journals/ai/HoffartSBW13-YAGO2} add temporal information to represent the KGs more accurately, and some event KGs~\cite{DVN/28117_2015-ICEWS2015,leetaru2013-gdelt} also contain the timestamps indicating when the events occurred. In the temporal knowledge graphs (TKGs), a fact is expanded into a quadruple $(e_h, r, e_t, \tau)$, where $\tau$ represents the timestamps.
\begin{figure}
  \centering  
  \includegraphics[width=0.98\linewidth]{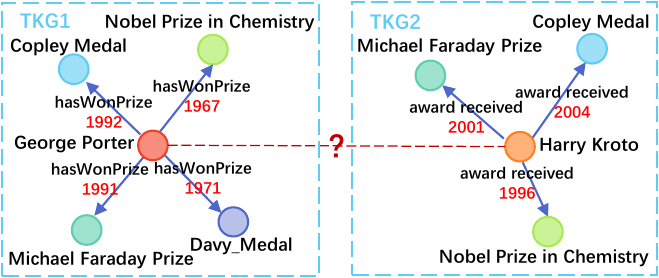}\\
  \caption{Example of the wrong alignment between similar entities in different temporal knowledge graphs.}
  \label{fig:example}
\end{figure}

Entity alignment (EA) seeks to find the same entities in the real world between different KGs, which is important for knowledge fusion between multi-source and multi-lingual KGs. In recent years, the embedding-based EA methods ~\cite{DBLP:journals/pvldb/SunZHWCAL20-EASurvey} have been widely investigated, which represent the entities in the low-dimensional vector space, and calculate the similarity of these vectors to obtain the equivalent entity pairs. Earlier EA methods~\cite{DBLP:conf/ijcai/ChenTYZ17-MTransE,DBLP:conf/semweb/SunHL17-JAPE} are based on the translation model. However, since the inability of such models to effectively capture graph structures, graph neural networks (GNN)-based models~\cite{DBLP:conf/emnlp/WangLLZ18-GCNAlign,DBLP:conf/wsdm/MaoWXLW20-MRAEA} emerge and achieve superior performance. Due to the scalability limitation of GNN~\cite{GNNBook2022}, these models are not suitable for large-scale KGs. LightEA~\cite{DBLP:conf/emnlp/MaoWWL22-LightEA} significantly enhances the efficiency of EA by utilizing a three-view label propagation instead of GNN.

Despite the success of the above EA methods, they all ignore the temporal information in the KGs, thus may lead to wrong alignment between similar entities in different KGs. Take Figure~\ref{fig:example} as an example. There are two subgraphs from YAGO and Wikidata. \emph{George Porter} in TKG1 and \emph{Harray Kroto} in TKG2 have similar structure and relations. Both of them have the same neighbors (\emph{Copley Medal}, \emph{Nobel Prize in Chemistry}, and \emph{Michael Faraday Prize}) and relations (\emph{hasWonPrize} in TKG1 is same as \emph{award recieved} in TKG2). \emph{George Porter} has another neighbor \emph{Davy Medal} and relation \emph{hasWonPrize}. Existing EA methods disregard temporal information of KGs and may wrongly take \emph{George Porter} and \emph{Harray Kroto} as an equivalent entity pair. 

Recently, time-aware methods~\cite{DBLP:conf/emnlp/XuS021-TEAGNN,DBLP:conf/www/XuSX022-TREA} emerge to improve the performance of EA between TKG. STEA~\cite{DBLP:conf/coling/CaiMMYZL22-STEA} adopts a simple GNN with a temporal information matching mechanism and achieves state-of-the-art (SOTA) performance. All of these TEA methods are based on GNN~\cite{GNNBook2022}, which has an inherent defect: the GNN is trained using the gradient descent algorithm and takes much time to converge to the optimal solution. Therefore, to promote the development of EA between TKGs, a straightforward approach is to combine the advantages of EA and TEA methods. 

To this end, we propose an effective and efficient non-neural EA framework between TKGs, namely LightTEA, which consists of four key components:
\emph{(1) Two-aspect Three-view Label Propagation.} We combine relational-aspect and temporal-aspect three-view label propagation (LP) to improve the performance of EA. In this module, GNN is replaced by LP, which does not require gradient propagation to train the neural networks, greatly reducing the time consumption of the model. 
\emph{(2) Sparse Similarity with Temporal Constraints.} Instead of calculating the similarity of all entities, we only retrieve the $top$-$k$ nearest neighbors of each entity to find the equivalent entity pairs, which saves the time complexity and space complexity of calculation. By utilizing the temporal information of each entity as the constraints of entity similarity, the EA performance is also improved. 
\emph{(3) Sinkhorn Operator.} To further promote the model's effectiveness, we regard the EA problem as a one-to-one assignment problem and use the Sinkhorn operator to solve it. The Sinkhorn operator is a fast and completely parallelizable algorithm and only takes seconds to converge to the optimal solution.
and \emph{(4) Temporal Iterative Learning.} Several studies have demonstrated that iterative learning effectively enhances EA and helps address the lack of alignment seeds in the real scenario. We adopt a temporal iterative learning strategy, which utilizes the additional temporal information of entities to get more credible alignment pairs for augmenting the training set to obtain better alignment results. 
In general, this paper presents the following contributions: 
\begin{itemize}
\item We propose an effective and efficient TEA framework that consists of four essential components: (1) Two-aspect Three-view Label Propagation, (2) Sparse Similarity with Temporal Constraints, (3) Sinkhorn Operator, and (4) Temporal Iterative Learning. All of these modules work together to improve the performance of EA while reducing time consumption. 

\item The proposed TEA framework combines the strengths of the latest SOTA EA and TEA models and addresses the limitations of current EA models that do not effectively utilize time information, as well as the scalability constraints of TEA models due to GNN usage. 

\item Extensive experiments on public datasets indicate that our proposed model significantly outperforms the SOTA methods for EA between TKGs, and the time consumed by LightTEA is only dozens of seconds at most, no more than 10$\%$ of the most efficient TEA method.
\end{itemize}

\begin{figure*}
  \centering  
  \includegraphics[width=0.99\linewidth]{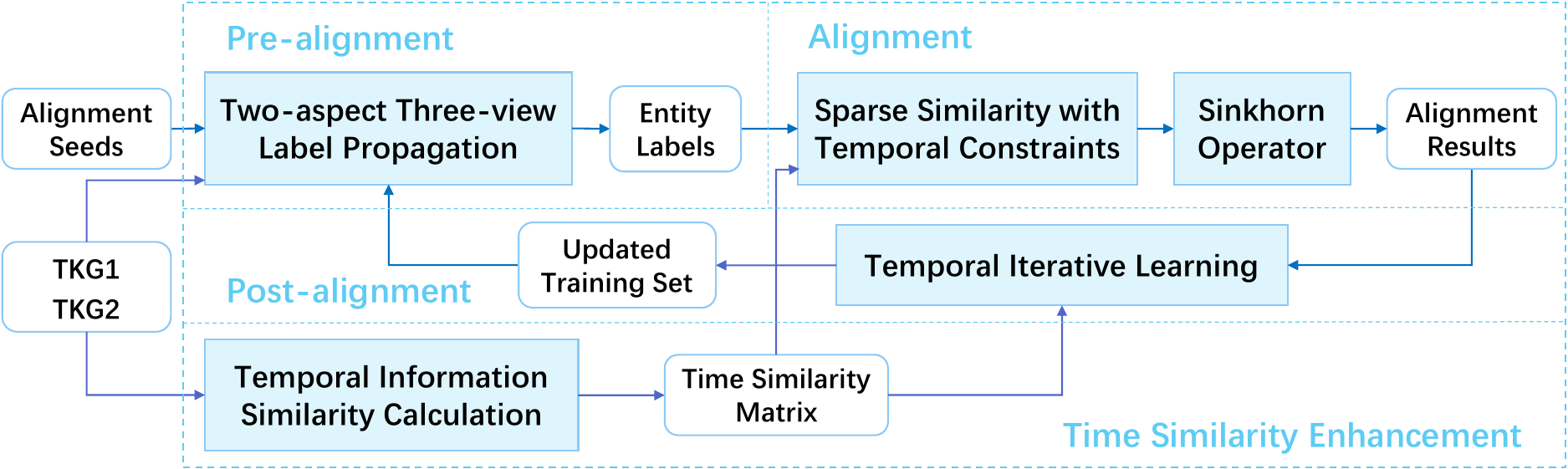}\\
  \caption{The framework of LightTEA. }
  \label{fig:framwork}
  \vspace{-1em}
\end{figure*}

\section{Related Work}
\subsection{Entity Alignment}
The purpose of EA is to find the equivalent entity pairs from different KGs. EA usually adopts embedding-based approaches, which are divided into two sub-categories: translation-based and GNN-based models.

Translation-based models regard the relations as the translation from the head entities to the tail entities, such as $(\bm{h}_{e_h} + \bm{h}_r \approx \bm{h}_{e_t})$. 
MTransE~\cite{DBLP:conf/ijcai/ChenTYZ17-MTransE} is the early entity alignment model based on TransE~\cite{DBLP:conf/nips/BordesUGWY13-TransE}, which maps two KGs into different vector spaces and considers the entities with similar positions in the two vector spaces as equivalent pairs. In addition to learning the structure embeddings of entities based on TransE with the relation triples, JAPE~\cite{DBLP:conf/semweb/SunHL17-JAPE} joins the attribute embedding and structure embedding of entities to align entities. BootEA~\cite{DBLP:conf/ijcai/SunHZQ18-BootEA} obtains the alignment-oriented KG embeddings and proposes a bootstrapping process by adding likely alignment entities into training data iteratively to improve the performance of EA.

GNN-based models promote EA by utilizing the graph structure of KGs. GCN-Align~\cite{DBLP:conf/emnlp/WangLLZ18-GCNAlign} encodes the entities into a unified vector space via GCN~\cite{DBLP:conf/iclr/KipfW17-GCN} and aligns the entities with their structure embeddings and attribute embeddings. However, GCN-Align does not effectively utilize the relation of KGs. MRAEA~\cite{DBLP:conf/wsdm/MaoWXLW20-MRAEA} learns the entity embeddings through a relation-aware self-attention GNN to obtain the alignment entities. RREA~\cite{DBLP:conf/cikm/MaoWXWL20-RREA} proposes a GNN with relational reflection to get the relation-specific embeddings and uses an iterative strategy to enhance EA.

The above methods focus on encoding the entity embeddings and aligning the entities by calculating the similarity of their embeddings, DATTI~\cite{DBLP:conf/acl/MaoMYZWXWL22-DATTI} applies a decoding process using the adjacency and inner correlation isomorphisms of KGs to the advanced methods and gains significant performance improvements. LightEA~\cite{DBLP:conf/emnlp/MaoWWL22-LightEA} adopts a three-view label propagation approach for entity alignment instead of using GNN and achieves comparable performance with SOTA EA methods while taking only one-tenth of the time consumption of those methods.

Although these methods have significantly advanced the development of EA, they all have the limitation of neglecting the temporal information in KGs.

\subsection{Time-aware Entity Alignment}
Recently, the research on TKGs has developed rapidly, and TEA methods have also sprung up.
TEA-GNN~\cite{DBLP:conf/emnlp/XuS021-TEAGNN} is the first TEA method using temporal information in KGs. It introduces a time-aware attention mechanism to learn entity embedding based on GNN and constructs five datasets extracted from ICEWS, YAGO3, and Wikidata to evaluate the model.
TREA~\cite{DBLP:conf/www/XuSX022-TREA} utilizes a temporal relational attention mechanism to integrate relational and temporal features of entities from the neighborhood and adopts the margin-based multi-class log-loss (MML) with sequential time regularizer to train the model. TREA is the most efficient TEA model since the MML can achieve fast convergence. STEA~\cite{DBLP:conf/coling/CaiMMYZL22-STEA} presents a simple GNN to learn the entity embeddings and uses a temporal information matching mechanism to calculate the time similarity of entities. Then it balances the time similarity and embedding similarity of entities to obtain the equivalent pairs.

By using the temporal information, the TEA methods achieve better performance. However, since the experimental datasets are much smaller than real-world KGs, these methods focus on improving performance and ignore efficiency. Although TREA utilizes MML to accelerate convergence, all TEA methods are based on GNN~\cite{GNNBook2022} and suffer from scalability issues. So we propose an effective and efficient TEA framework to address the limitations.

\section{Problem Formulation}
A TKG can be formalized as $G=(E, R, T, Q)$, where $E$, $R$ and $T$ are the sets of entities, relations, and timestamps respectively, $Q\subset E\times R\times E\times T$ denotes the set of quadruples. A quadruple stores the real-world fact and can be presented as $(e_h,r,e_t,\tau)$, where $e_h,e_t \in E$. Given two TKGs $G_1=(E_1,R_1,T_1,Q_1)$, $G_2=(E_2,R_2,T_2,Q_2)$, and alignment seeds set $S=\{(e_{1_i},e_{2_j})|e_{1_i}\in E_1,e_{2_j}\in E_2, e_{1_i} \equiv e_{2_j} \}$ where $\equiv$ denotes equivalence. EA task aims to find new equivalent entity pairs between $G_1$ and $G_2$ based on $S$. $C$ is the set of reference entity pairs used for evaluation. Specifically, a uniform time set $T^* = T_1 \cup T_2$ is constructed by merging the timestamps in the two time sets. Therefore, the two TKGs can be renewed as $G_1=(E_1,R_1,T^*,Q_1)$ and $G_2=(E_2,R_2,T^*,Q_2)$ sharing the same set of timestamps.

\section{The Proposed Approach}
The LightTEA framework can be described as three phases with time similarity enhancement. (1) For \emph{pre-alignment} phase, we use two-aspect three-view label propagation to learn the labels of entities. (2) For \emph{alignment} phase, we first compute the sparse similarity with temporal constraints, then translate the EA problem to the assignment problem and utilize the Sinkhorn operator to resolve it. (3) For \emph{post-alignment} phase, we adopt a temporal iterative learning strategy to enhance EA. Figure~\ref{fig:framwork} shows the framework of LightTEA.

\subsection{Two-aspect Three-view Label Propagation}

\begin{figure*}
  \centering  
  \includegraphics[width=0.99\linewidth]{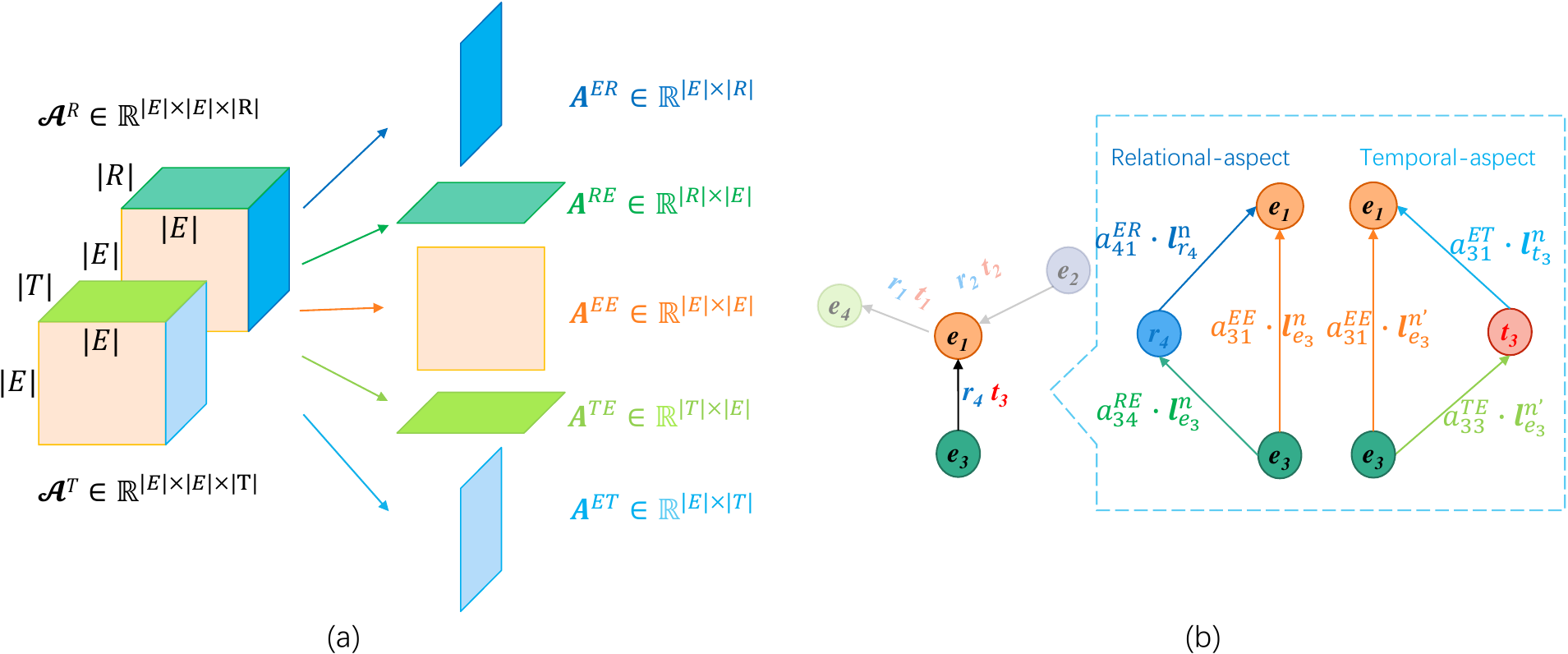}\\
  \caption{Illustrations of Two-aspect Three-view Label propagation. a) Regarding TKG as two three-order tensors in terms of relational and temporal aspects, which include five types of adjacency relationships. b) Label propagation from $e_3$ to $e_1$ using the two-aspect three-view approach.}
  \label{fig:2-aspectLP}
  \vspace{-1em}
\end{figure*}

Inspired by LightEA~\cite{DBLP:conf/emnlp/MaoWWL22-LightEA}, we extend the three-view label propagation to two-aspect (relational-aspect and temporal-aspect) three-view label propagation, which can enhance EA with temporal information while not increase the time and space complexity.

Specifically, the TKG requires a four-order tensor $\boldsymbol{\mathscr{A}} \in \mathbb{R}^{|E| \times |E| \times |R| \times |T|}$ to fully describe the adjacency relations. As shown in Figure~\ref{fig:2-aspectLP}(a), we regard TKG as two three-order tensor $ \boldsymbol{\mathcal{A}}^R \in \mathbb{R}^{|E| \times |E| \times |R|} $ and $ \boldsymbol{\mathcal{A}}^T \in \mathbb{R}^{|E| \times |E| \times |T|}$, there are five-view in TKGs: $\boldsymbol{A}^{ER} \in \mathbb{R}^{|E| \times |R|}$, $\boldsymbol{A}^{RE} \in \mathbb{R}^{|R| \times |E|}$, $\boldsymbol{A}^{EE} \in \mathbb{R}^{|E| \times |E|}$, $\boldsymbol{A}^{TE} \in \mathbb{R}^{|T| \times |E|}$, and $\boldsymbol{A}^{ET} \in \mathbb{R}^{|E| \times |T|}$, which represent the adjacency relations from head entity to relation, relation to tail entity, head entity to tail entity, timestamps to tail entity, head entity to timestamps, respectively. Then, we use the relational-aspect and temporal-aspect three-views label propagation to update the labels of entities, relations, and timestamps (as shown in Figure~\ref{fig:2-aspectLP}(b)). 

The relational-aspect three-views label propagation can be presented as follows:
\begin{align}
    &\boldsymbol{L}_e^{(n+1)} = \boldsymbol{A}^{EE} \cdot \boldsymbol{L}_e^{(n)} + \boldsymbol{A}^{ER} \cdot \boldsymbol{L}_r^{(n)} \\
    &\boldsymbol{L}_r^{(n+1)} = \boldsymbol{A}^{RE} \cdot \boldsymbol{L}_e^{(n)}
    \label{eq1&2}
\end{align}
where $\boldsymbol{L}_e \in \mathbb{R}^{|E| \times |d|}$, $\boldsymbol{L}_r \in \mathbb{R}^{|R| \times |d|}$ are label matrixs of entities and relations, ${\cdot}$ means the dot product. Following LightEA, we regard each pair of alignment entities as an independent class, and independently sample random vectors on the $d$-dimensional hyper-sphere to approximate the one-hot label vectors for representing the alignment seeds, so we set $\boldsymbol{l}_{e_i}^{(0)} = \boldsymbol{l}_{e_j}^{(0)} = random(d)\; \forall{(e_i,e_j)} \in S$. For the entities in the candidate set $C$, we initialize the label of them as $\boldsymbol{l}_{e_k}^{(0)} = \boldsymbol{l}_{e_l}^{(0)} = 0\; \forall{(e_k,e_l)} \in C$, and the initial label matrix of relations is also set to all-zero $\boldsymbol{L}_r^{(0)}=0$. 
 
 The final label of entity $e_i$ in the relational aspect is the concatenation of label vectors in all steps:
\begin{equation}
    \boldsymbol{l}_{e_i}^{r} = [\boldsymbol{l}_{e_i}^{(0)}||\boldsymbol{l}_{e_i}^{(1)}||...||\boldsymbol{l}_{e_i}^{(n-1)}||\boldsymbol{l}_{e_i}^{(n)}]
    \label{eq3}
\end{equation}

The temporal-aspect three-views label propagation can be expressed as follows:
\begin{align}
    &\boldsymbol{L}_e^{(n+1)'} = \boldsymbol{A}^{EE} \cdot \boldsymbol{L}_e^{(n)'} + \boldsymbol{A}^{ET} \cdot \boldsymbol{L}_t^{(n)}\\
    &\boldsymbol{L}_t^{(n+1)} = \boldsymbol{A}^{TE} \cdot \boldsymbol{L}_e^{(n)'}
    \label{eq4&5}
\end{align}
where $\boldsymbol{L}_e^{'} \in \mathbb{R}^{|E| \times |d|}$, and the initializtion label matrix of $\boldsymbol{L}_e^{(0)'}$ is the same as $\boldsymbol{L}_e^{(0)}$. $\boldsymbol{L}_t \in \mathbb{R}^{|T| \times |d|}$ is label matrix of timestamps and is initialized to $\boldsymbol{L}_t^{(0)}=0$. 

We concatenate the label vectors of all steps as the final label of entity $e_i$ in the temporal aspect:
\begin{equation}
    \boldsymbol{l}_{e_i}^{t} = [\boldsymbol{l}_{e_i}^{(0)'}||\boldsymbol{l}_{e_i}^{(1)'}||...||\boldsymbol{l}_{e_i}^{(n-1)'}||\boldsymbol{l}_{e_i}^{(n)'}]
    \label{eq6}
\end{equation}

Finally, the label of entity $e_i$ is the balanced result of the two aspect labels:
\begin{equation}
    \boldsymbol{l}_{e_i} = (1-\alpha) \times \boldsymbol{l}_{e_i}^{r} + \alpha \times \boldsymbol{l}_{e_i}^{t}
    \label{eq7}
\end{equation}
where $\alpha$ is a hyper-parameter to balance the label of relational aspect and temporal aspect.

\begin{table*}
\centering
\resizebox{\textwidth}{!}{
\begin{tabular}{lrrrrrrrrr}
\toprule
{Datasets} & {$|E_1|$} & {$|E_2|$} & {$|R_1|$} & {$|R_2|$} & {$|T^*|$} & {$|Q_1|$} & {$|Q_2|$} & {$|S|$} & {$|C|$}\\
\midrule
\textbf{DICEWS-1K/200} & {9,517} & {9,537} & {247} & {246} & {4,017} & {307,552} & {307,553} & {1,000/200} & {7,566/8,366} \\
\textbf{YAGO-WIKI50K-5K/1K} & {49,629} & {49,222} & {11} & {30} & {245} & {221,050} & {317,814} & {5,000/1,000} & {44,172/48,172}\\
\bottomrule
\end{tabular}}
\caption{Statistics of DICEWS-1K/200 and YAGO-WIKI50K-5K/1K. $|\cdot|$ denotes the numbers.}
\label{tab:datasets}
\end{table*}

\subsection{Temporal Information Similarity Calculation}
Recent research(~\cite{DBLP:conf/coling/CaiMMYZL22-STEA}) suggests that the temporal information similarity can enhance the EA between TKGs, so we calculate the time similarity matrix and use it in the alignment and post-alignment phase.

First, we collect all timestamps of entities. Then we calculate the time similarity $s^t_{e_ie_j}$ of $e_i$ and $e_j$ by the following: 
\begin{equation}
    s^t_{e_ie_j} = \frac {2 \times v} { k + q } 
    \label{eq8}
\end{equation}
where $v$ denote the number of same items of $e_i$ and $e_j$, $k$ and $q$ are the numbers of timestamps of $e_i$ and $e_j$, respectively. 

\subsection{Sparse Similarity with Temporal Constraints}
After obtaining the labels of entities, we calculate the similarity of entities by their labels in a sparse way with temporal constraints. 

Early studies~\cite{DBLP:conf/ijcai/ChenTYZ17-MTransE,DBLP:conf/emnlp/WangLLZ18-GCNAlign,DBLP:conf/emnlp/XuS021-TEAGNN} calculate the embedding similarity (Cosine, Euclidean, or Manhattan distance) of all entities to find the equivalent entity pairs. The calculation complexity is $O(|E|^2 d)$, and the space complexity of the similarity matrix is $O(|E|^2)$. LightEA~\cite{DBLP:conf/emnlp/MaoWWL22-LightEA} notices that the similarity of many entities is very small and infinitely close to zero. Even if these smaller values are removed initially, it does not significantly affect the alignment results. Therefore, instead of calculating the similarities between all entities, we only retrieve the $top$-$k$ nearest neighbors for each entity by approximate nearest neighbor (ANN) algorithms~\footnote{In LightTEA, we use the FAISS framework~\cite{DBLP:journals/tbd/JohnsonDJ21-FAISS} for approximate vector retrieval }. It only takes several seconds to find the $top$-$k$ nearest neighbors, and the space complexity of the sparse similarity matrix is $O(|E| k), k \ll |E|$.

Different from LightEA, we use the time similarity matrix as a constraint to obtain the sparse similarity of entities. Specifically, we get the sparse similarity matrix $S^{l} \in \mathbb{R}^{|E| \times k}$ by the ANN algorithms with the labels of entities and find the related time similarity of these entities, the final sparse similarity of entities is as follows:
\begin{equation}
   \boldsymbol{S}' = (1- \beta) \times \boldsymbol{S}^{l} + \beta \times \boldsymbol{S}^{t}
    \label{eq9}
\end{equation}
where $\beta$ is the hyper-parameter for balancing the label similarity and time similarity of entities.

\subsection{Sinkhorn Operator} \label{sec:Sinkhorn}
Existing TEA methods~\cite{DBLP:conf/emnlp/XuS021-TEAGNN,DBLP:conf/www/XuSX022-TREA,DBLP:conf/coling/CaiMMYZL22-STEA} 
simply calculate the similarity of the entities to obtain the equivalent entity pairs in the alignment phase. To further improve the effectiveness of the model, we adopt the Sinkhorn operator in the alignment phase to enhance EA. 

LightEA~\cite{DBLP:conf/emnlp/MaoWWL22-LightEA} regards the EA problem as a one-to-one assignment problem to improve the performance of EA. The goal of the assignment problem is to find the optimal strategy to obtain the maximum profit and can be formulated as follows:
\begin{equation}
    \underset{P \in \mathcal{P}_{|E|}}{\arg \; \max} \; {\langle \boldsymbol{P}, \, \boldsymbol{S} \rangle _F}
    \label{eq10}
\end{equation}
where $\boldsymbol{P}$ is a permutation matrix that has exactly one entry of 1 in each row and each column and 0s elsewhere. $\mathcal{P}_{|E|}$ is the set of $|E|\times |E|$ permutation matrices, $\boldsymbol{S} \in \mathbb{R}^{|E| \times |E|}$ is the similarity matrix of entities, and $\langle \cdot \rangle _F$ represents the Frobenius inner product.

The Sinkhorn operator proposes a fast and completely parallelizable algorithm for the assignment problem. It iteratively normalizes rows and columns of the similarity matrix:
\begin{equation}
\begin{aligned}
    Sinkhorn^0(\boldsymbol{S}) &= exp(\boldsymbol{S}), \\
    Sinkhorn^m(\boldsymbol{S}) &= \mathcal{N}_c(\mathcal{N}_r(Sinkhorn^{m-1}(\boldsymbol{S}))), \\
    Sinkhorn(\boldsymbol{S}) &= \underset{m \rightarrow \infty}{\lim } Sinkhorn^m(\boldsymbol{S})
    \label{eq11}
\end{aligned}
\end{equation}
where $\mathcal{N}_r(\boldsymbol{S}) = \boldsymbol{S} \oslash (\boldsymbol{S}1_N1_N^{\top})$, $\mathcal{N}_c(\boldsymbol{S}) = \boldsymbol{S} \oslash (1_N1_N^{\top}\boldsymbol{S})$ are the row and column-wise normalization operators, $\oslash$ denotes the element-wise division, $1_N$ represents a column vector of ones, and $m$ is the iterations. The time complexity of Sinkhorn is $O(m|E|^2)$.

We also regard the EA problem as an assignment problem and employ the Sinkhorn operator to obtain the approximate solution:
\begin{equation}
\begin{aligned}
    & \underset{\boldsymbol{P} \in \mathcal{P}_{|E|}}{\arg \; \max} \; {\langle \boldsymbol{P}, \, \boldsymbol{S}' \rangle _F} \\
    & = \underset{t \rightarrow 0^+}{lim} \; Sinkhorn(\boldsymbol{S}'/ \, t)
   \label{eq12}
\end{aligned}
\end{equation}
where $\boldsymbol{S}' \in \mathbb{R}^{|E| \times k}$ is the sparse similarity matrix with temporal constraints, which is calculated by equation~\eqref{eq9}, $t$ is the temperature. In this way, the performance of EA improves significantly, and the computational complexity drops to $O(m|E|k)$.

\subsection{Temporal Iterative Learning}
Iterative Learning is proposed by BootEA~\cite{DBLP:conf/ijcai/SunHZQ18-BootEA} to address the problem of fewer alignment seeds and enhance the performance of EA. It is also called a semi-supervised alignment strategy which continuously selects possible entity pairs to augment the training data through an iterative method in the post-alignment phase. 

To promote EA, we adopt temporal iterative learning. Different from STEA~\cite{DBLP:conf/coling/CaiMMYZL22-STEA}, which adopts a bi-directional iterative strategy, we simply choose the entity and its nearest neighbor whose similarity value is greater than the threshold~\footnote{The threshold is $0.8$ in LightTEA.} as the alignment pairs and add them to the training set for the next iteration.

\begin{table*}
\centering
\resizebox{\textwidth}{!}{
\begin{tabular}{llrrrrrrrrrrrr}
\toprule
 & & \multicolumn{3}{c}{\bf DICEWS-1K} & \multicolumn{3}{c}{\bf DICEWS-200} & \multicolumn{3}{c}{\bf YAGO-WIKI50K-5K} &  \multicolumn{3}{c}{\bf YAGO-WIKI50K-1K}\\
\cmidrule(rl){3-5}%绘制第2列和第4列的横线，留空
\cmidrule(rl){6-8}
\cmidrule(rl){9-11}
\cmidrule(rl){12-14}
Categories & Methods & {MRR} & {Hits@1} &  {Hits@10} & {MRR} & {Hits@1} &  {Hits@10} & {MRR} & {Hits@1} &  {Hits@10} & {MRR} & {Hits@1} &  {Hits@10}\\
\midrule
{Supervised} & {MTransE} & {.150} & {.101} & {.241} & {.104} & {.067} & {.175} & {.332} & {.242} & {.477} & {.033} & {.012} & {.067}\\
 & {JAPE} & {.198} & {.144} & {.298} & {.138} & {.098} & {.210} & {.345} & {.271} & {.488} & {.157} & {.101} & {.262}\\
 & {AlignE} & {.593} & {.508} & {.751} & {.303} & {.222} & {.457} & {.800} & {.756} & {.883} & {.618} & {.565} & {.714} \\
 & {GCN-Align} & {.291} & {.204} & {.466} & {.231} & {.165} & {.363} & {.581} & {.512} & {.711} & {.279} & {.217} & {.398}\\
 & {MRAEA} & {.745} & {.675} & {.870} & {.564} & {.476} & {.733} & {.848} & {.806} & {.913} & {.685} & {.623} & {.801}\\
 & {LightEA*} & {.833} & {.785} & {.918} & {.779} & {.721} & {.878} & {.960} & {.948} & {.979} & {.902} & {.878} & {.945}\\
 \cmidrule{2-14}
 & {TEA-GNN} & {.911} & {.887} & {.947} & {.902} & {.876} & {.941} & {.909} & {.879} & {.961} & {.775} & {.723} & {.871}\\
 & {TREA} & {.933} & {.914} & {.966} & {.927} & {.910} & {.960} & {.958} & {.940} & {.989} & {.885} & {.840} & {.937}\\
 & {STEA*} & {.941} & {.928} & {.960} & {.941} & {.927} & {.961}  & {.954} & {.935} & {.986} & {.916} & {.887} & {.966}\\
 \cmidrule{2-14}
 & {LightTEA*} & {\underline{.959}} & {\underline{.952}} & {\underline{.970}} & {\underline{.955}} & {\underline{.949}} & {.966}  & {\underline{.990}} & {\underline{.986}} & {\underline{.997}} & {\underline{.977}} & {\underline{.969}} & {\underline{.989}}\\
  & {$p$-value} & {2e-7} & {3e-7} & {2e-6} & {5e-7} & {1e-7} & {5e-5} & {6e-9} & {1e-9} & {7e-7} & {2e-9} & {5e-10} & {4e-8}\\
 \midrule
{Semi-supervised} & {BootEA} & {.670} & {.598} & {.796} & {.614} & {.546} & {.737} & {-} & {-} & {-} & {-} & {-} & {-} \\
& {RREA} & {.840} & {.795} & {.917} & {.823} & {.773} & {.911} & {.913} & {.887} & {.955} & {.870} & {.836} & {.929}\\
& {LightEA} & {.875} & {.838} & {.936} & {.878} & {.842} & {.937} & {.963} & {.951} & {.980} & {.951} & {.938} & {.970}\\
\cmidrule{2-14}
 & {STEA} & {.954} & {.945} & {.967} & {.954} & {.943} & {\underline{.968}} & {.974} & {.961} & {.992} & {.962} & {.943} & {\underline{.989}}\\
 \cmidrule{2-14}
 & {LightTEA} & {\bf.960} & {\bf.954} & {\bf.970} & {\bf.963} & {\bf.958} & {\bf.972}  & {\bf.992} & {\bf.988} & {\bf.997} & {\bf.992} & {\bf.988} & {\bf.997}\\
 & {$p$-value} & {2e-4} & {2e-5} & {4e-4} & {4e-6} & {2e-6} & {5e-5} & {2e-7} & {2e-8} & {5e-5} & {1e-8} & {2e-9} & {2e-6}\\
\bottomrule
\end{tabular}}
\caption{ Experimental results on all datasets. - means the results are not obtained. The best results are written in bold. Underline indicate the sub-optimal results. The $p$-value is the result of one sample t-test between LightTEA*/LightTEA and their corresponding strong baselines.}
\label{tab:results}
\end{table*}

\section{Experiments}
We conduct the experiments on a workstation with a GeForce RTX 3090 GPU and an AMD EPYC 7502 32-Core Processor CPU, 128GB memory. The codes and datasets will be available on GitHub~\footnote{https://github.com/lcai2/LightTEA}.

\subsection{Datasets}
To comprehensively evaluate the effectiveness and efficiency of the proposed model, we experiment on two widely used public datasets. The statistics of these datasets are listed in Table~\ref{tab:datasets}.

(1) \textbf{DICEWS-1K/200}~\cite{DBLP:conf/emnlp/XuS021-TEAGNN} is constructed from the event knowledge graph ICEWS05-15 which contains events during 2005 to 2015. It consists of two subsets with different alignment seeds 1K/200. 

(2) \textbf{YAGO-WIKI50K-5K/1K}~\cite{DBLP:conf/emnlp/XuS021-TEAGNN} extracts the equivalent entities with temporal information from YAGO and Wikidata. There are two subsets, one with 5K alignment seeds and the other with 1K.
 
\subsection{Baselines}
In the experiments, we compared our proposed model with two categories of advanced entity alignment methods:

(1) \textbf{Supervised methods}:  JAPE~\cite{DBLP:conf/semweb/SunHL17-JAPE}, AlignE~\cite{DBLP:conf/ijcai/SunHZQ18-BootEA}, GCN-Align~\cite{DBLP:conf/emnlp/WangLLZ18-GCNAlign}, MRAEA~\cite{DBLP:conf/wsdm/MaoWXLW20-MRAEA}, LightEA*~\cite{DBLP:conf/emnlp/MaoWWL22-LightEA}, TEA-GNN~\cite{DBLP:conf/emnlp/XuS021-TEAGNN}, TREA~\cite{DBLP:conf/www/XuSX022-TREA}, and STEA*~\cite{DBLP:conf/coling/CaiMMYZL22-STEA}. JAPE, AlignE, GCN-Align, MRAEA, and LightEA* are EA methods, TEA-GNN, TREA, and STEA* are TEA methods.

(2) \textbf{Semi-supervised methods}: BootEA~\cite{DBLP:conf/ijcai/SunHZQ18-BootEA}, RREA~\cite{DBLP:conf/cikm/MaoWXWL20-RREA}, LightEA~\cite{DBLP:conf/emnlp/MaoWWL22-LightEA}, STEA~\cite{DBLP:conf/coling/CaiMMYZL22-STEA}. BootEA, RREA, and LightEA are EA methods, STEA is a TEA method.

The main experiment results of these methods reported in the paper are from STEA (SOTA TEA method), except the latest SOTA EA method LightEA. The experiments of LigthEA are implemented on its open-source code. For a fair comparison, LightTEA has two corresponding versions: (1) LightTEA* is the supervised version. (2) LightTEA is the semi-supervised version with iterative learning. 

\subsection{Settings}

\subsubsection{Evaluation Metrics}
Following the conventions, we adopt mean reciprocal rank (MRR) and Hits@$k$ $(k=1,10)$ as the evaluation metrics. MRR reports the average reciprocal of the ranks, and Hits@$k$ calculates the proportion of correct alignment pairs whose rank is not greater than $k$. In particular, Hits@1 represents the accuracy of the results, which is the most important indicator. The higher the MRR and Hits@$k$, the better the performance.

\subsubsection{Implementation Details}
We use the fixed training set and validation set provided by TEA-GNN. The hyper-parameters are set as follows: the dimension of the hyper-sphere is $d=512$, the factor $\alpha$ is set to $ 0.6$ for DICEWS and 
$0.5$ for YAGO-WIKI50K to balance the relational-aspect and temporal-aspect label propagation, the factor $\beta$ is set to $0.4$ for balancing the label similarity and time similarity. Following LightEA, we use rounds $n=2$ for two-aspect three-view label propagation, retrieve the $top$-$k=500$ nearest neighbors for sparse similarity, set the number of iterations of the Sinkhorn operator to $m=15$, and set the temperature to 
$t=0.05$. The reported performances are the averages of five independent runs.

\subsection{Main Experiments}
Table~\ref{tab:results} lists the main experimental results of our proposed model and all baselines on the four datasets. Among the supervised methods, LightTEA* has significant improvements in all metrics. Compared to the SOTA supervised TEA method STEA*, LightTEA* improves the Hits@1 by 2.61\%, 2.33\%, 5.45\%, and 9.22\%, respectively. In the semi-supervised methods, LightTEA outperforms all the baselines on all datasets across all metrics. The improvements of Hits@1 compared with STEA are 0.93\%, 1.57\%, 2.79\%, and 4.77\%, respectively. TEA methods perform better in these two categories of methods than EA methods by using temporal information. The semi-supervised methods achieve better performance than supervised methods, which indicates the effectiveness of temporal iterative learning. Without iteration, LightTEA* outperforms all baselines except for STEA's Hits@10 on DICEWS-200 with tiny gap $0.002$. The high performance of LightTEA* demonstrates that the two-aspect three-view label propagation, the sparse similarity with temporal constraints, and the Sinkhorn operator are effective in promoting EA.
We conduct one sample t-test between LightTEA(/LightTEA*) and their corresponding strong baselines. All the $p$-value $<< 0.01$ indicates that our proposed model significantly outperforms all baselines.

\subsection{Ablation Study}
We conduct an ablation experiment on DICEWS-200 and YAGO-WIKI50K-1K to investigate the contributions of three critical components of LightTEA: (1) Temporal-aspect Label Propagation (TLP), (2) Temporal Constraints (TC), and (3) Sinkhorn Operator (SO). 

\begin{table} [ht]
\centering
\resizebox{\linewidth}{!}{
\begin{tabular}{lrrrrrr}
 \toprule
 & \multicolumn{3}{c}{\bf DICEWS-200} & \multicolumn{3}{c}{\bf YAGO-WIKI50K-1K}\\
\cmidrule(rl){2-4}%绘制第2列和第4列的横线，留空
\cmidrule(rl){5-7}
{Methods} & {MRR} & {Hits@1} &  {Hits@10} & {MRR} & {Hits@1} &  {Hits@10}\\
 \midrule
{LightTEA} & {\bf.963} & {\bf.958} & {\bf.972} & {\bf.992} & {\bf.988} & {\bf.997}\\
{ w/o TLP} & {.956} & {.950} & {.965} & {.989} & {.985} & {.994}\\
{ w/o TC} & {.954} & {.947} & {.965} & {.987} & {.982} & {.995}\\
{ w/o SO} & {.942} & {.933} & {.954} & {.962} & {.947} & {.986}\\
\bottomrule
\end{tabular}}
\caption{Ablation study of LightTEA on DICEWS-200 and YAGO-WIKI50K-1K. w/o means without.}
\label{tab:ablation}
\end{table}

As reported in Table~\ref{tab:ablation}, without temporal-aspect label propagation (w/o TLP), the performance of LightTEA drops a little, which indicates the effectiveness of the temporal-aspect label propagation. When removing the temporal constraints (w/o TC), the underperformance of LightTEA implies the effectiveness of the temporal constraints. Without the Sinkhorn operator (w/o SO), the performance of LightTEA drops significantly, demonstrating the contribution of utilizing the Sinkhorn operator in the alignment phase to enhance EA.

\subsection{Efficiency Studys}
Table~\ref{tab:timecosts} shows the overall time costs on DICEWS-200 and YAGO-WIKI50K-1K datasets by TEA methods from data loading to evaluation. The results of TREA are from~\cite{DBLP:conf/www/XuSX022-TREA} since it doesn't provide the source codes. The other results are obtained by directly running the source codes provided by the author. 

\begin{table}
\centering
\resizebox{\linewidth}{!}{
\begin{tabular}{llrr}
 \toprule
{Categories} & {Methods} & {\bf DICEWS-200} & {\bf YAGO-WIKI50K-1K}\\
 \midrule
{supervised} & {TEA-GNN} & {1,547} & {5,042} \\ 
 & {TREA} & {128} & {2,655} \\
 & {LightTEA*} & {5} & {24} \\
 \midrule
{semi-supervised} & {STEA} & {578} & {5,428} \\ 
 & {LightTEA} & {12} & {85} \\
\bottomrule
\end{tabular}}
\caption{Time costs of TEA methods on DICEWS-200 and YAGO-WIKI50K-1K (seconds).}
\label{tab:timecosts}
\end{table}

It can be seen from Table~\ref{tab:timecosts}, in the supervised time-aware methods, TREA costs less time than TEA-GNN with MML, which speed up the convergence. LightTEA* takes much less time than the above two methods. The time costs of LigthTEA* are 5 seconds and 24 seconds on the two datasets, respectively, which are only $3.91 \%$ and $0.90 \%$ of TREA. In the semi-supervised methods, the time costs of LightTEA are 12 seconds and 85 seconds, which are $2.08 \%$ and $1.70 \%$ of STEA on the two datasets, respectively. The time consumed by LightTEA is no more than $10 \%$ of the most efficient methods TREA. By using temporal iterative learning, LightTEA increases the time consumption while improving performance compared to LightTEA*. The high efficiency of LightTEA indicates that it could be applied to large-scale datasets for EA between TKGs.

\subsection{Hyper-parameter Analysis}

We conduct experiments on the following hyper-parameters to investigate their effect on the performance of LightTEA. 

(1) The dimension of hyper-sphere $d$. We select the dimension in the set $\{64, 128, 256, 512, 1024, 2048\}$ and conduct the experiments. The Hits@1 scores with different dimensions on DICEWS-200 are shown in Table~\ref{tab:dims}.

\begin{table}
\centering
\resizebox{\linewidth}{!}{
\begin{tabular}{lrrrrrr}
 \toprule
\#\ Dimension & 64 & 128 & 256 & \textbf{512} & 1024 & 2048 \\
 \midrule
LightTEA* & 0.943 & 0.947 & 0.947 & \textbf{0.949} & 0.949 & 0.949 \\
LightTEA & 0.948 & 0.954 & 0.956 & \textbf{0.958} & 0.958 & 0.958 \\
\bottomrule
\end{tabular}}
\caption{Hits@1 scores with different dimensions on DICEWS-200.}
\label{tab:dims}
\end{table}

From the table we can see that as the dimension increases, the Hits@1 score for both LightTEA and LightTEA* gradually improves until it reaches their best performance at the dimension of 512. Even if the dimension is further increased, the performance remains unchanged. This indicates that when the dimension of the label vector is greater than 512, it only increases memory consumption and provides no benefit to performance.

\begin{figure}
  \centering  
  \includegraphics[width=0.99\linewidth]{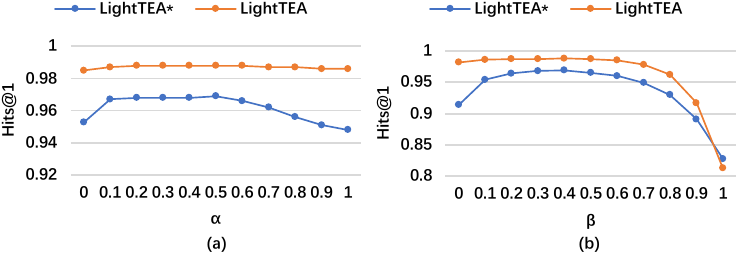}\\
  \caption{Experimental results with different $\alpha$ and $\beta$ on YAGO-WIKI50K-1K.}
  \label{fig:alphabeta}
\end{figure}

(2) The balance factors $\alpha$ and $\beta$. We set the two factors in range 0$\sim$1 with interval 0.1 to investigate the impact of different values. The experimental results with different $\alpha$ and $\beta$ on YAGO-WIKI50K-1K are shown in Figure~\ref{fig:alphabeta}.

$\alpha$ balances the relational-aspect label propagation (RLP) and temporal-aspect label propagation (TLP). Figure~\ref{fig:alphabeta}(a) shows the Hits@1 of LightTEA* and LightTEA with different $\alpha$. Due to the use of temporal iterative learning, the performance of LightTEA does not change significantly with different $\alpha$. The Hits@1 curve of LightTEA* shows that the result of combining RLP and TLP with appropriate weight ($\alpha = 0.5$) is better than using only one of them ($\alpha = 0$ or $\alpha = 1$).

$\beta$ is used to balance the label similarity and time similarity. As shown in Figure~\ref{fig:alphabeta}(b), with the increase of $\beta$, the Hits@1 first increases slowly (reaching the maximum value when $\beta = 0.4$), and then decreases rapidly. When $\beta = 1$, Hits@1 of LightTEA with temporal iterative learning are lower than LightTEA*, it indicates that only using time similarity to generate the possible alignment pairs and add them to the training set for the next iteration will hurt the performance of the model.

\section{Conclusion}
Existing EA methods ignore the temporal information in KGs, which may wrongly align similar entities. The TEA methods lack scalability since the inherent defect of GNN. To address the limitations, we proposes an effective and efficient TEA framework that consists of four important components: (1) Two-aspect Three-view Label Propagation, (2) Sparse Similarity with Temporal Constraints, (3) Sinkhorn Operator, and (4) Temporal Iterative Learning. These modules work collaboratively to enhance the model's performance while reducing time consumption.

Extensive experiments on public datasets indicate that the proposed model significantly outperforms the SOTA methods. The time consumed by the model is only dozens of seconds at most, no more than 10$\%$ of the most efficient TEA methods, which denotes that our model has high efficiency and can be applied to large-scale TKGs.

\section*{Acknowledgements}
We appreciate the support from National Natural Science Foundation of China with the Main Research Project on Machine Behavior and Human-Machine Collaborated Decision Making Methodology (72192820 \& 72192824), Pudong New Area Science \& Technology Development Fund (PKX2021-R05), Science and Technology Commission of Shanghai Municipality (22DZ2229004), and Shanghai Trusted Industry Internet Software Collaborative Innovation Center. 

%% The file named.bst is a bibliography style file for BibTeX 0.99c
\bibliographystyle{named}
\bibliography{ijcai23-LightTEA}

\end{document}